%% file: main.tex
\newif\ifcomments  
\newif\ifsupp  
\commentstrue
\suppfalse

\documentclass[a4paper,table]{article}
\usepackage{INTERSPEECH2022}
\usepackage[dvipsnames]{xcolor}
\usepackage{highlight}
\usepackage{comment}
\usepackage{hyperref}
\usepackage{url}
\usepackage{graphicx}

\newcommand{\ot}[1]{\ifcomments\textcolor{green}{\textbf{OT: #1}}\else \ignorespaces \fi}

\title{Detecting Unintended Memorization in Language-Model-Fused ASR}
\name{W. Ronny Huang, Steve Chien, Om Thakkar, Rajiv Mathews}

\address{Google LLC}
\email{\{wrh, schien, omthkkr, mathews\}@google.com}

\input{pkgs}

\input{macros}

\begin{document}
\maketitle
\begin{abstract}
End-to-end (E2E) models are often being accompanied by language models (LMs) via shallow fusion for boosting their overall quality as well as recognition of rare words. 
At the same time, several prior works show that LMs are susceptible to unintentionally memorizing rare or unique sequences in the training data. 
In this work, we design a framework for detecting memorization of random textual sequences (which we call canaries) in the LM training data when one has only black-box (query) access to LM-fused speech recognizer, as opposed to direct access to the LM.
On a production-grade Conformer RNN-T E2E model fused with a Transformer LM,  we show that detecting memorization of singly-occurring canaries from the LM training data of 300M examples is possible.
Motivated to protect privacy, we also show that such memorization gets significantly reduced by per-example gradient-clipped LM training without compromising overall quality.
\end{abstract}


\input{1_intro}

\input{2_rel_work}

\input{3_method}
\input{4_setup}
\input{5_expts}

\input{6_concl}

\bibliographystyle{IEEEtran}
\bibliography{references}

\end{document}

%% file: pkgs.tex
\usepackage{amsmath,amssymb,pifont}
\usepackage{multicol}
\usepackage{amstext}
\usepackage{amsthm}
\usepackage{multirow}
\usepackage{booktabs}
\usepackage[skip=0pt]{subcaption}
\usepackage{times}
\usepackage{lipsum}
\usepackage[shortlabels]{enumitem}
\usepackage{cancel}
\usepackage{wrapfig}
\usepackage{array}
\usepackage{siunitx}
\usepackage{csvsimple}
\usepackage[multidot]{grffile}
\usepackage{bbm}
\usepackage{dblfloatfix}
\usepackage{hyperref}
\usepackage{makecell}
\usepackage{bbm, dsfont}
\usepackage{mathtools}
\usepackage{xcolor}
\usepackage{comment}

\usepackage[multiple]{footmisc}
\usepackage{mathrsfs}
\usepackage{tikz}
\usepackage{hyperref}
\usepackage{cleveref}

\usepackage{algpseudocode,algorithm,algorithmicx}
\usepackage{xfrac}

%% file: macros.tex
\hypersetup{final}

\renewcommand{\vec}[1]{#1}

\makeatletter
\newcommand{\vast}{\bBigg@{4}}
\newcommand{\Vast}{\bBigg@{5}}
\makeatother

\newcommand{\ex}[2]{{\ifx&#1& \mathbb{E} \else
\underset{#1}{\mathbb{E}} \fi \left[#2\right]}}
\newcommand{\pr}[2]{{\ifx&#1& \mathbb{P} \else
\underset{#1}{\mathbb{P}} \fi \left[#2\right]}}

\DeclarePairedDelimiterX{\ip}[2]{\langle}{\rangle}{#1, #2}

\DeclarePairedDelimiterX{\infdivx}[2]{(}{)}{%
  #1\;\delimsize\|\;#2%
}

\newcommand{\mypar}[1]{\smallskip
	\noindent{\textbf{{#1}:}}}
	
\renewcommand{\epsilon}{\varepsilon}

%% file: 1_intro.tex
\section{Introduction}
\label{sec:intro}

Neural networks for common ASR applications like smart assistants, dictation, and telephony, are trained on millions of utterances which can often contain sensitive information, such as credit card numbers.
Given the prevalence of ASR in user products, ensuring the privacy of training data is an increasingly important topic in the speech community \cite{DTR21, DTR221, sp_id}.

Modern E2E models dedicate most of their resources to encoding audio. 
For example, the Conformer(L) architecture \cite{gulati2020conformer} dedicates 17 layers to the encoder to process audio, and only 1 layer to the decoder to process text. This may appear to limit an E2E model's ability to
memorize its textual data.
However, the ASR pipeline is often equipped with an accompanying language model (LM)
whose job is to use its linguistic knowledge to disambiguate acoustically similar hypotheses given by the acoustic model (AM)\footnote{In this paper, we refer to an E2E model as an AM to better contrast it from an LM. We acknowledge the term ``acoustic model'' can historically refer to only the audio processing component, e.g.,~an encoder. 
}.
As such, LMs are often integrated into production speech engines, improving WER by 5-10\% relative~\cite{zhang2020pushing} with larger gains on rare words~\cite{huang2022lookup}.

For better domain match, such LMs are often trained on the transcripts of the same data as the AM.
Additionally, personalized or contextually biased \cite{aleksic2015bringing} ASR systems use a standard AM with an LM trained on per-user information (e.g. contacts list \cite{michaely2016unsupervised}, typing patterns \cite{sim2019investigation}, etc.)---packing yet more personal information into the LM.
Several prior works~\cite{CLEKS19, SS19, TRMB20, RTM20, CTWJHLRBS21, CIJLTZ} have shown that LMs are susceptible to unintentionally memorizing rare or unique sequences of data, thus making textual training data susceptible to such leakage.
This leaves the open question, which we try to answer in this work: {\em Does adding an LM to an AM create a privacy vulnerability in the combined ASR system even without direct access to the LM?}

Our first contribution is to design a framework to detect unintended memorization using only query access to the ASR system. We do this by inserting a set of random ``canaries'' into the LM's training data, and observing the ASR model's performance on these examples. We need to overcome a pair of obstacles: First, since ASR models accept audio and not text, we use a Text-To-Speech (TTS) system to generate synthetic utterances for querying the ASR system. Next, unlike conventional attacks on LMs which require some confidence score, e.g.,~perplexity, ASR systems typically only give the single top transcript for a given audio sequence. Thus, we use the model's WER on canaries versus baseline examples as our primary metric.

We test for detecting memorization across a range of canary frequencies and LM capacities. Our results show that memorization is detectable even for canaries that appear only once in a large (300M sample) training corpus, and then grows significantly the more frequently a canary is repeated in the training data (from 1 to 32 times). We also show how increasing model capacity leads to increased memorization.

Our next contribution is to study the effect of per-example gradient clipping on mitigating memorization, an idea motivated by prior work~\cite{CLEKS19, TRMB20}. Our experiments here demonstrate that this technique can significantly reduce the level of memorization without appreciable loss of overall utility in the model.

Finally, we observe that while using \emph{clean} TTS utterances can be useful for detecting memorization, there can be instances where the ASR system may not leverage the LM for providing the top transcripts, e.g.,~if the TTS utterances are   easy to predict for the AM itself.
In such cases, the LM's memorization may not be able to surface to the top transcript predictions which we use for inference.
Thus, we add noise to parts of the TTS utterances such that the LM's predictions can be leveraged by the ASR system.
Moreover, we design a simple classifier that can use the top transcripts for these utterances to conduct a \emph{membership inference} (MI) task.
We provide an analysis of how metrics like precision and recall (commonly used for evaluating MI attacks~\cite{shokri2017membership, yeometal, CCNSTT}) can be used to demonstrate the extent of unintended memorization in ASR LMs.
For instance, we show in our experiments that while our classifier shows a baseline (50\% precision, 13\% recall) for non-memorized sequences of data, it achieves ($\sim$80\% precision, $\sim$50\% recall) for sequences that appeared only eight times among 300M other samples for LM training and were subsequently memorized.

%% file: 2_rel_work.tex
\mypar{Related Work}
 Recent work has shown that model updates during ASR training can leak potentially sensitive artifacts like labels~\cite{DTR21} and speaker identity~\cite{DTR221} of utterances used in computing the updates. 
There have been  works~\cite{CLEKS19, SS19, TRMB20, RTM20, CTWJHLRBS21, CIJLTZ} that focus on designing extraction techniques to demonstrate the susceptibility of LMs to memorize training data.
There are also works~\cite{F19, BBFST21} that theoretically study such memorization and how it relates to generalization in learning.

%% file: 3_method.tex
\section{Method}
\label{sec:method}

\subsection{Detecting unintended memorization}
\label{sec:framework}
We start by defining a (publicly-known) distribution over the tokens of the model. 
Next, we draw some i.i.d.~samples from the distribution and add these ``canaries” with varying frequencies into the LM's normal training set. Simultaneously, we also draw a equal-sized set of ``extraneous'' samples in the same manner, which are not used for training. 

After training the LM, we query an AM+LM recognizer to detect the level the unintended memorization of the canaries. This is done by first using a WaveNet TTS system~\cite{tts} to generate synthetic utterances for the canaries and extraneous examples. These are fed into the recognizer, and the top-1 transcripts are retrieved.
We then use two methods to measure unintended memorization: First, we compare the word error rate (WER) of the canaries from a canary-trained model, and contrast it with the same measurement from an extraneous-trained model. Second,
as described later in \S\ref{sec:classifier}, we also design a membership classifier that modifies the TTS utterances to accurately predict memorization on a \textit{per-example} basis.

\subsection{Training using per-example gradient clipping}
\label{sec:clipping}
An established technique for reducing unintended memorization in ML models is DP-SGD~\cite{BST14, DP-DL}.
This algorithm modifies standard gradient descent in two ways: it first clips the gradient of each training example to a fixed maximum {\em clip norm}, and then adds random noise to the mean clipped gradient. 
A model trained this way is {\em differentially private}~\cite{DMNS}, which is a formal notion for quantifying privacy leakage. 
Unfortunately, large models trained using DP-SGD with strong privacy guarantees tend to have unacceptably low utility (i.e. generalization) compared with baseline (non-private) models.

A possible step forward is to only perform  per-example clipping, thus bounding the \emph{sensitivity} of each gradient.
A model trained this way no longer satisfies any meaningful differential privacy, but prior work~\cite{CLEKS19, TRMB20} has shown that such models tend to match the utility of baseline models while being empirically less prone to memorization.


\subsection{Membership classifier}
\label{sec:classifier}
A conventional membership inference attack uses a membership classifier that predicts whether or not a candidate example is in the training set based on the true label and the output posteriors of the model for that candidate example.
In our case, there are two challenges.
First, realistic speech recognizers only output the top-1 hypothesis;
thus, this is the only artifact available to the membership classifier.
This situation is akin to label-only membership inference attacks~\cite{choquette2021label}.
Second, the model we wish to probe here is the LM, but we can only send inputs to the AM.
Thus, we must synthesize TTS audio of the candidate text sequence for which we wish to predict the membership.
This presents its own set of challenges.
On one hand, if the TTS is too clear, the AM alone might predict the sequence perfectly without the LM's help.
On the other hand, if the TTS is too unclear, the AM could output an empty transcript, inhibiting the LM's ability to influence the prediction.

We opt for a middle ground:
The first part of the candidate example, which we call the prefix, is synthesized with clear TTS,
while the remainder, which we call the suffix, has Gaussian noise added. 
This partially obscured example is then
fed into the fused model, and its output observed. From here, we use a 
very simple membership classifier: if the model's output exactly matches the ground truth, the classifier predicts that the example was used in training (i.e.~is a canary); otherwise, the classifier predicts the example was not used in training.

%% file: 4_setup.tex
\section{Setup}

\subsection{Canaries}
\label{sec:canarysetup}
Each canary is a fixed-length sequence of  random lower-case alphabetical characters with spaces between them, e.g. ``\texttt{o e g d b u}''.
This format can mimic some sensitive data like passwords, serial numbers, secret codes, etc.
Since our goal is to detect unintended memorization, such a distinct format helps avoid artifacts due to learning from the training data.
Further, this format has a well-defined baseline likelihood of $26^{-n}$, where 26 is the number of possible characters, and $n$ is the sequence length.
This is useful for confirming the extent to which the LM has memorized the canary sequences by comparing their perplexity to $26^n$,
which we do during training.
For each canary, we insert it into the LM training data at a certain frequency (i.e.,~number of times it's repeated). 
We choose frequencies that are powers of 2, as shown in Table \ref{tab:counts}.
For e.g., there are 16384 canaries that appear once in the training set (CAN1), 8192 canaries that appear twice (CAN2), and so on until CAN32.
We also keep a set of canaries not inserted into the training set (CAN0) to use as a baseline during evaluations.

\begin{table}[h!]
\vspace{-5pt}
    \small
    \caption{Canary datasets. Each dataset is inserted into training at the frequency indicated.}
    \vspace{-10pt}
    \label{tab:counts}
    \centering
    \smallskip\noindent
    \resizebox{1.0\columnwidth}{!}{%
    \begin{tabular}{lccccccc}
        \toprule
        Name &  CAN0 & CAN1     & CAN2     & CAN4     & CAN8     & CAN16    & CAN32    \\
        Frequency &     0 & 1     & 2     & 4     & 8     & 16    & 32    \\
        \midrule
        \# Canary     & 16384 & 16384 & 8192  & 4096  & 2048  & 1024  & 512      \\
        \# Extraneous  &  16384 & 16384 & 8192  & 4096  & 2048  & 1024  & 512       \\
        \bottomrule
    \end{tabular}%
    }
\end{table}

We also craft another equally-sized set of \textit{extraneous} sequences that take the same structure and the same frequency distribution as the canaries.
The only difference is that these sequences are sampled independently, and contain no matches with the canaries.
The extraneous set is used for training a baseline LM that has a similar distribution as the LM trained with canaries.
In addition, the extraneous set is used to compute the precision and recall metrics discussed in \S\ref{sec:mia}.


\subsection{Dataset}
\label{sec:datasetup}
Our main ASR task is voice search (VS); our test set consists of 11k utterances from Google's VS traffic. Each utterance is only a few words long.
The training set for both the AM and LM is a slice of Google search traffic from multiple domains such as voice search, farfield, telephony, and YouTube, making up $\sim$300M utterances. 
All utterances are anonymized and hand-transcribed, with the exception of YouTube being semi-supervised.
Our data collection abides by Google AI Principles \cite{aiprinciples}.
This is the same training set used in Google's production speech \cite{sainath2020streaming} and language \cite{huang2022sentence} models.
For the speech model, the acoustics are further diversified via multi-style training \cite{kim2017mtr}, random down-sampling from 16 to 8 kHz, and SpecAug \cite{Park2019}.
The LM uses only the transcripts of the training set.
The canaries are merged into the LM's training set (the AM is not trained on canaries), and make up less than 0.1\% of the total.

\subsection{Architecture}
Our acoustic model is the same as in \cite{sainath2021cascadedlm}. It is a state-of-the-art Conformer-encoder, streaming RNN-T model with 150M parameters emitting 4096 lower-case wordpieces.
The decoder is a tiny 2M-parameter stateless embedding decoder \cite{Rami21} with no recurrence.
Our language model is a 70M-parameter, 10-layer transformer \cite{Vaswani17} with model dimension of 768, feed-forward dimension of 3072, and 5 attention heads. It is trained using the AdaFactor optimizer with a batch size of 1024 for 800k steps. 

\subsection{Language model fusion}
Shallow fusion \cite{kannan2018analysis}, i.e.~interpolation of the AM and LM logits during decoding time, is the predominant LM fusion method used in performance speech recognition and is the method we use here.
In particular, we apply HAT shallow fusion which includes an additional interpolation term that factorizes out the AM's log-posterior from its internal language model score \cite{variani2020hybrid}.
In summary, the decoder optimizes
\begin{equation*}
    y^* = \underset{y}{\mathrm{argmax}}\; \left[\log p(y|x) - \lambda_2 \log p_{\rm ILM}(y) + \lambda_1 p_{\rm LM}(y)\right],
\end{equation*}
\noindent
where $p(y|x)$ is the AM log-posterior score, $p_{\rm ILM}(y)$ is the AM's internal language model score, and $p_{\rm LM}$ is the LM score.
The interpolation weights ($\lambda_1$, $\lambda_2$) are optimized via Vizier \cite{golovin2017google} where the objective is to reduce WER on the VS test set.

%% file: 5_expts.tex
\section{Experiments}
\label{sec:expt}

\subsection{Baseline comparisons}
Table~\ref{tab:lmsize} shows the performance of a collection of recognizers, each consisting of a different LM fused with the same AM.
Since the AM is fixed for all experiments, we name each row by the LM for brevity;
note that each row represents the decoded WER results from an entire AM+LM recognizer.
For each recognizer, we decode on the VS test set to measure overall recognizer quality.
We also decode on the TTS of the single-frequency canaries, CAN1, as well as CAN0 as a baseline.
Recall \S\ref{sec:canarysetup} and Table \ref{tab:counts} for discussion on these canary datasets.

\begin{table}[h!]
\centering
\caption{WER on VS, CAN0, and CAN1 for different LMs.
Lower values indicate more generalization (VS) or memorization (CAN0 or CAN1).
Naming convention for LMs is \{size\}-\{trained\_on\_canary\_or\_extraneous\_set\}.}
\label{tab:lmsize}
\resizebox{0.63\columnwidth}{!}{%
\begin{tabular}{l|ccc}
\toprule
LM             & VS   & CAN0  & CAN1  \\
\midrule
B1: None           & 6.53 & 34.02 & 33.98 \\
B2: 70M            & 6.22 & 33.88 & 34.50 \\
B3: 70M-EXT        & 6.24 & 23.59 & 24.21 \\
\midrule
E1: 9M-CAN         & 6.51 & 20.93 & 21.20 \\
E2: 30M-CAN        & 6.25 & 22.13 & 21.43 \\
E3: 70M-CAN        & 6.26 & 22.97 & 21.17 \\
E4: 130M-CAN       & 6.22 & 23.46 & 20.77 \\
\bottomrule
\end{tabular}%
}
\end{table}

We first introduce several baselines.
B1 is the AM alone without an accompanying LM, and B2 is the AM fused with a 70M parameter LM trained on the same data.
B2 provides a 5\% WER relative improvement on the VS test set over B1, confirming that LM shallow fusion is properly configured to provide overall gains.
B3's LM is also 70M parameters but trained on the extraneous (EXT) dataset; thus it has learned the same \textit{structure} as the canaries (6 space-separated alphabetical letters), but not their specific \textit{sequences}.
As a result, B3's CAN0 and CAN1 WERs are 10\% lower than that of B1 and B2, which were not trained on the same structure.
B3 is the most appropriate baseline for comparing our canary-trained LM results as it factors out the memorization of canary \textit{sequence} from the learning of canary \textit{structure}.

\begin{figure}[ht!]
\vspace{-0.2cm}
\centering
\includegraphics[width=0.85\columnwidth]{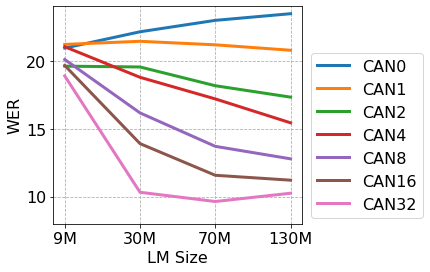}
\vspace{-0.2cm}
\caption{WER on canary sets of various frequencies as a function of LM size. Lower values indicate more memorization.}\label{fig:lmsize}
\end{figure}


\subsection{Unintended memorization}
\label{sec:memoresults}
Experiments E1-E4 in Table \ref{tab:lmsize} consist of different-sized LMs (9M, 30M, 70M, 130M) trained on the canaries.
Each size was obtained simply by applying a constant multiplier to the number of attention heads, model dimension, and feed-forward-layer dimension of the Transformer architecture.
As a sanity check, training on canaries does not significantly compromise overall performance as indicated by the fact that VS WER is comparable to B2 and B3.
The exception is E1, which provides no gain to VS due to its small LM size.

On CAN0 (canaries not in training) the WER increases as the model size increases (E1-E4).
This is explained by the fact that larger LMs have greater modeling power, and stronger LMs are better able to assign high perplexities to random sequences they have not seen in training.
On the other hand, larger LMs are also better able to assign low perplexities to sequences they \textit{have} seen in training.
This pattern is evident for single-frequency canaries in the CAN1 column where WER decreases with larger models.
In Figure \ref{fig:lmsize}, we plot the WER versus LM size dependence for all canary frequencies.
As expected, the trend of increased memorization (lower WER) with LM size is even stronger for more frequent canaries (CAN2-CAN32).
Thus, while increasing LM model size is a common way to improve performance, unfortunately this has the side-effect of making unintended memorization worse.




\subsection{Mitigation via per-example gradient clipping}
Now that we have shown that LMs can memorize even single frequency canaries in a large dataset, we focus on a strategy to mitigate the memorization.
In Table~\ref{tab:dpresults}, we report the results on different levels of per-example gradient clipping (discussed in \S\ref{sec:clipping}).
At a clip norm of 32, no gradients are clipped. Thus, the top row can be seen as a non-clipped baseline that is most prone to memorization, similar to the models in \S\ref{sec:memoresults}.
At a clip norm of 0.5, 99\% of the per-example gradients get clipped.
At each clip norm level, we train an LM on the canaries (70M-CAN), along with a corresponding LM on the extraneous set (70M-EXT) as a baseline.
This allows us to make well-baselined measurements of canary memorization, as
the only difference between the WER given by 70M-CAN and 70M-EXT is due to the memorization of the canary \textit{sequences}.
All other factors, including the canary \textit{structure} and clip norm level, are identical between 70M-CAN and 70M-EXT for each row.

\begin{table}[]
\centering
\caption{WER of VS and CAN1 at different levels of gradient clipping.
Each testset is evaluated on 70M-CAN and 70M-EXT.
WERR is WER relative of 70M-CAN vs. 70M-EXT.
Less negative values of WERR indicate more memorization mitigation.
}
\label{tab:dpresults}
\resizebox{\columnwidth}{!}{%
\begin{tabular}{l|cc|ccc}
\toprule
Clip      & \multicolumn{2}{c}{VS}   & \multicolumn{3}{c}{CAN1}  \\
norm      & 70M-EXT & 70M-CAN  & 70M-EXT & 70M-CAN & WERR (\%)  \\
\midrule
32        & 6.24    & 6.26     & 24.21   & 20.45   & -15.5 \\
16        & 6.23    & 6.30     & 25.16   & 22.55   & -10.4 \\
8         & 6.27    & 6.22     & 25.12   & 23.10   & -8.0  \\
4         & 6.27    & 6.29     & 24.79   & 23.48   & -5.3  \\
2         & 6.26    & 6.29     & 25.00   & 23.39   & -6.4  \\
1         & 6.30    & 6.23     & 24.93   & 23.37   & -6.3  \\
0.5       & 6.19    & 6.28     & 24.39   & 23.50   & \textbf{-3.6}  \\
\bottomrule
\end{tabular}%
}
\end{table}

A first observation is that the VS WER is mostly unchanged regardless of the clip norm level,
indicating that gradient clipping does not compromise model quality.
Next, on CAN1, 70M-CAN obtains lower WERs compared to the corresponding 70M-EXT baseline at each clip level due to memorization.
The amount of memorization can be measured as the WER relative of 70M-CAN with respect to 70M-EXT, and we display this value in the last column (WERR).
WERR starts at -15.5\% at a clip norm of 32 and reduces to -3.6\% at a clip norm of 0.5, making clear that the models trained with more aggressive clipping (lower clip norm) can achieve significantly less memorization.

\begin{figure}[]
\vspace{-0.2cm}
\centering
\includegraphics[width=0.9\columnwidth]{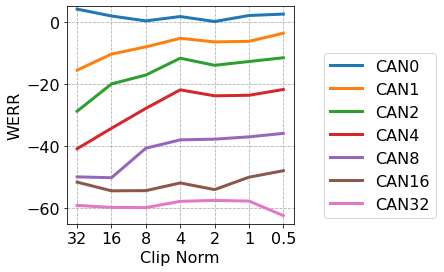}
\vspace{-0.4cm}
\caption{Percent WER relative (70M-CAN vs. 70M-EXT), or WERR, on canary sets as a function of the gradient clipping level. 
More negative WERR indicates more memorization. Conversely, less negative WERR indicates stronger mitigation.
}\label{fig:clip_wer}
\end{figure}

Figure \ref{fig:clip_wer} plots the WERR for canaries of all frequencies (CAN0-CAN32) with respect to clip norm.
Up to CAN8 (frequency of 8), there is a trend of memorization decreasing (WERR becoming less negative) with more aggressive clipping (smaller clip norm).
Beyond that (CAN16 and CAN32), canaries are so frequent that they might be considered data to be learned rather than idiosyncratic examples.
We see clipping has limited ability to mitigate memorization at these levels.



\subsection{Membership inference attacks}
\label{sec:mia}

We now examine whether the general reduction in memorization (as measured by WER) induced by gradient clipping in the previous section can be used to identify individual training examples via a membership inference attack. We use the classifier described previously in \S\ref{sec:classifier}. Recall that this classifier works by taking the partially obscured TTS audio waveform of a character sequence into the recognizer, and outputting ``yes'' if and only if the recognizer is exactly correct. 

We test our classifier on three models: (1) the model 70M-CAN described in Table~\ref{tab:lmsize}, whose training data includes the canaries but not the extraneous examples, and whose gradients are not clipped, (2) a clipped version of 70M-CAN in which gradients were clipped to norm 1, and (3) a baseline model whose training data does not include either the canaries or extraneous examples. Our main metrics for attack efficacy are precision and recall of the classifier over the combined canary and extraneous test sets. More explicitly, precision is the likelihood that a positive membership prediction is correct, and recall is the likelihood that each canary will be predicted as such. 

\begin{figure}[ht!]
    \centering
    \vspace{-1em}
    \includegraphics[width=0.7\columnwidth]{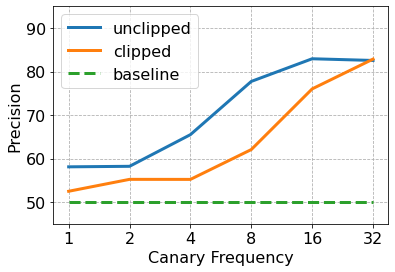}
    \includegraphics[width=0.7\columnwidth]{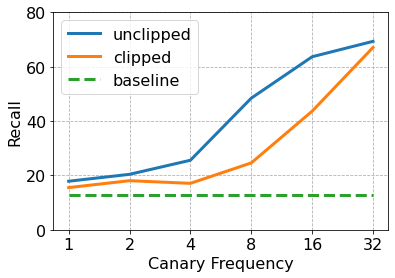}
    \caption{Precision (top) and recall (bottom) of the membership classifier on an unclipped model and a clipped model, as compared with a baseline model. The unclipped model (blue) shows memorization (precision$>$50\%) even for canaries with low frequency in the data. By comparison, the clipped model (orange) is able to mitigate memorization at low frequencies.}
    \label{fig:precision-recall}
\end{figure}

Figure~\ref{fig:precision-recall} shows our results. The classifier applied to the baseline model has 50\% precision and 13\% recall. The precision value is equivalent to random guessing since the baseline model has seen neither the canary or extraneous datasets. Given the nature of our classifier, the recall value indicates the fraction of examples the baseline model is able to perfectly recognize, and serves as a reference value for the other two models.

Compared with the baseline model, the unclipped model shows substantially improved precision and recall, even on canaries that appear only once in the data (58\% precision and 18\% recall), and rising quickly from there as the
canary frequency increases. The membership classifier success saturates once canary frequency reaches 16, with precision above 80\% and recall close to 70\%. By comparison, our results show that the clipped model is more robust against such memorization---both precision and recall remain relatively low until canary frequency reaches 8, and the attack does not saturate until a frequency of 32. Taken as whole, these results confirm that the results from \S\ref{sec:memoresults} do in fact carry over to membership inference.


%% file: 6_concl.tex
\section{Conclusion}
In this work, we designed a framework for detecting unintended memorization of rare or unique textual data in E2E ASR systems with fused LMs.
We demonstrated the extent to which such memorization of random sequences of data can be detected using our framework by conducting experiments on a production-grade Conformer RNN-T ASR model with a Transformer LM.
Lastly, we also showed the extent to which this gets reduced using per-example gradient clipping.

%% file: main.bbl
\begin{thebibliography}{10}
\providecommand{\url}[1]{#1}
\csname url@samestyle\endcsname
\providecommand{\newblock}{\relax}
\providecommand{\bibinfo}[2]{#2}
\providecommand{\BIBentrySTDinterwordspacing}{\spaceskip=0pt\relax}
\providecommand{\BIBentryALTinterwordstretchfactor}{4}
\providecommand{\BIBentryALTinterwordspacing}{\spaceskip=\fontdimen2\font plus
\BIBentryALTinterwordstretchfactor\fontdimen3\font minus
  \fontdimen4\font\relax}
\providecommand{\BIBforeignlanguage}[2]{{%
\expandafter\ifx\csname l@#1\endcsname\relax
\typeout{** WARNING: IEEEtran.bst: No hyphenation pattern has been}%
\typeout{** loaded for the language `#1'. Using the pattern for}%
\typeout{** the default language instead.}%
\else
\language=\csname l@#1\endcsname
\fi
#2}}
\providecommand{\BIBdecl}{\relax}
\BIBdecl

\bibitem{DTR21}
\BIBentryALTinterwordspacing
T.~Dang, O.~Thakkar, S.~Ramaswamy, R.~Mathews, P.~Chin, and F.~Beaufays,
  ``Revealing and protecting labels in distributed training,'' \emph{CoRR},
  vol. abs/2111.00556, 2021. [Online]. Available:
  \url{https://arxiv.org/abs/2111.00556}
\BIBentrySTDinterwordspacing

\bibitem{DTR221}
\BIBentryALTinterwordspacing
------, ``A method to reveal speaker identity in distributed {ASR} training,
  and how to counter it,'' \emph{CoRR}, vol. abs/2104.07815, 2021. [Online].
  Available: \url{https://arxiv.org/abs/2104.07815}
\BIBentrySTDinterwordspacing

\bibitem{sp_id}
\BIBentryALTinterwordspacing
A.~S. Shamsabadi, B.~M.~L. Srivastava, A.~Bellet, N.~Vauquier, E.~Vincent,
  M.~Maouche, M.~Tommasi, and N.~Papernot, ``Differentially private speaker
  anonymization,'' 2022. [Online]. Available:
  \url{https://arxiv.org/abs/2202.11823}
\BIBentrySTDinterwordspacing

\bibitem{gulati2020conformer}
A.~Gulati, J.~Qin, C.-C. Chiu, N.~Parmar, Y.~Zhang, J.~Yu, W.~Han, S.~Wang,
  Z.~Zhang, Y.~Wu \emph{et~al.}, ``Conformer: Convolution-augmented transformer
  for speech recognition,'' \emph{Proc. Interspeech 2020}, pp. 5036--5040,
  2020.

\bibitem{zhang2020pushing}
Y.~Zhang, J.~Qin, D.~S. Park, W.~Han, C.-C. Chiu, R.~Pang, Q.~V. Le, and Y.~Wu,
  ``Pushing the limits of semi-supervised learning for automatic speech
  recognition,'' \emph{arXiv preprint arXiv:2010.10504}.

\bibitem{huang2022lookup}
W.~R. Huang, T.~N. Sainath, C.~Peyser, S.~Kumar, D.~Rybach, and T.~Strohman,
  ``Lookup-table recurrent language models for long tail speech recognition,''
  \emph{arXiv preprint arXiv:2104.04552}, 2021.

\bibitem{aleksic2015bringing}
P.~Aleksic, M.~Ghodsi, A.~Michaely, C.~Allauzen, K.~Hall, B.~Roark, D.~Rybach,
  and P.~Moreno, ``Bringing contextual information to google speech
  recognition,'' 2015.

\bibitem{michaely2016unsupervised}
A.~H. Michaely, M.~Ghodsi, Z.~Wu, J.~Scheiner, and P.~Aleksic, ``Unsupervised
  context learning for speech recognition,'' in \emph{2016 IEEE Spoken Language
  Technology Workshop (SLT)}.

\bibitem{sim2019investigation}
K.~C. Sim, P.~Zadrazil, and F.~Beaufays, ``An investigation into on-device
  personalization of end-to-end automatic speech recognition models,''
  \emph{arXiv preprint arXiv:1909.06678}, 2019.

\bibitem{CLEKS19}
N.~Carlini, C.~Liu, {\'{U}}.~Erlingsson, J.~Kos, and D.~Song, ``The secret
  sharer: Evaluating and testing unintended memorization in neural networks,''
  in \emph{28th {USENIX} Security Symposium, {USENIX} Security 2019, Santa
  Clara, CA, USA}, 2019.

\bibitem{SS19}
C.~Song and V.~Shmatikov, ``Auditing data provenance in text-generation
  models,'' in \emph{Proceedings of the 25th {ACM} {SIGKDD} International
  Conference on Knowledge Discovery {\&} Data Mining, {KDD} 2019, Anchorage,
  AK, USA}, 2019.

\bibitem{TRMB20}
\BIBentryALTinterwordspacing
O.~Thakkar, S.~Ramaswamy, R.~Mathews, and F.~Beaufays, ``Understanding
  unintended memorization in federated learning,'' \emph{CoRR}, vol.
  abs/2006.07490, 2020. [Online]. Available:
  \url{https://arxiv.org/abs/2006.07490}
\BIBentrySTDinterwordspacing

\bibitem{RTM20}
\BIBentryALTinterwordspacing
S.~Ramaswamy, O.~Thakkar, R.~Mathews, G.~Andrew, H.~B. McMahan, and
  F.~Beaufays, ``Training production language models without memorizing user
  data,'' \emph{CoRR}, vol. abs/2009.10031, 2020. [Online]. Available:
  \url{https://arxiv.org/abs/2009.10031}
\BIBentrySTDinterwordspacing

\bibitem{CTWJHLRBS21}
N.~Carlini, F.~Tram{\`{e}}r, E.~Wallace, M.~Jagielski, A.~Herbert{-}Voss,
  K.~Lee, A.~Roberts, T.~B. Brown, D.~Song, {\'{U}}.~Erlingsson, A.~Oprea, and
  C.~Raffel, ``Extracting training data from large language models,'' in
  \emph{30th {USENIX} Security Symposium, {USENIX} Security 2021}, 2021.

\bibitem{CIJLTZ}
\BIBentryALTinterwordspacing
N.~Carlini, D.~Ippolito, M.~Jagielski, K.~Lee, F.~Tram{\`{e}}r, and C.~Zhang,
  ``Quantifying memorization across neural language models,'' \emph{CoRR}, vol.
  abs/2202.07646, 2022. [Online]. Available:
  \url{https://arxiv.org/abs/2202.07646}
\BIBentrySTDinterwordspacing

\bibitem{shokri2017membership}
R.~Shokri, M.~Stronati, C.~Song, and V.~Shmatikov, ``Membership inference
  attacks against machine learning models,'' in \emph{2017 IEEE symposium on
  security and privacy (SP)}.\hskip 1em plus 0.5em minus 0.4em\relax IEEE,
  2017, pp. 3--18.

\bibitem{yeometal}
S.~Yeom, I.~Giacomelli, M.~Fredrikson, and S.~Jha, ``Privacy risk in machine
  learning: Analyzing the connection to overfitting,'' in \emph{2018 IEEE 31st
  Computer Security Foundations Symposium (CSF)}, Jul. 2018, pp. 268--282.

\bibitem{CCNSTT}
\BIBentryALTinterwordspacing
N.~Carlini, S.~Chien, M.~Nasr, S.~Song, A.~Terzis, and F.~Tram{\`{e}}r,
  ``Membership inference attacks from first principles,'' \emph{CoRR}, vol.
  abs/2112.03570, 2021. [Online]. Available:
  \url{https://arxiv.org/abs/2112.03570}
\BIBentrySTDinterwordspacing

\bibitem{F19}
\BIBentryALTinterwordspacing
V.~Feldman, ``Does learning require memorization? {A} short tale about a long
  tail,'' \emph{CoRR}, vol. abs/1906.05271, 2019. [Online]. Available:
  \url{http://arxiv.org/abs/1906.05271}
\BIBentrySTDinterwordspacing

\bibitem{BBFST21}
G.~Brown, M.~Bun, V.~Feldman, A.~D. Smith, and K.~Talwar, ``When is
  memorization of irrelevant training data necessary for high-accuracy
  learning?'' in \emph{{STOC} '21: 53rd Annual {ACM} {SIGACT} Symposium on
  Theory of Computing}, 2021.

\bibitem{tts}
\BIBentryALTinterwordspacing
A.~van~den Oord, Y.~Li, I.~Babuschkin \emph{et~al.}, ``Parallel wavenet: Fast
  high-fidelity speech synthesis,'' in \emph{Proceedings of the 35th
  International Conference on Machine Learning, {ICML} 2018,
  Stockholmsm{\"{a}}ssan, Stockholm, Sweden}, ser. Proceedings of Machine
  Learning Research, vol.~80. [Online]. Available:
  \url{http://proceedings.mlr.press/v80/oord18a.html}
\BIBentrySTDinterwordspacing

\bibitem{BST14}
R.~Bassily, A.~Smith, and A.~Thakurta, ``Private empirical risk minimization:
  Efficient algorithms and tight error bounds,'' in \emph{Proc. of the 2014
  IEEE 55th Annual Symp. on Foundations of Computer Science (FOCS)}, 2014, pp.
  464--473.

\bibitem{DP-DL}
M.~Abadi, A.~Chu, I.~J. Goodfellow, H.~B. McMahan, I.~Mironov, K.~Talwar, and
  L.~Zhang, ``Deep learning with differential privacy,'' in \emph{Proc. of the
  2016 {ACM} {SIGSAC} Conf. on Computer and Communications Security
  ({CCS}'16)}, 2016, pp. 308--318.

\bibitem{DMNS}
\BIBentryALTinterwordspacing
C.~Dwork, F.~McSherry, K.~Nissim, and A.~Smith, ``Calibrating noise to
  sensitivity in private data analysis,'' in \emph{Proc. of the Third Conf. on
  Theory of Cryptography (TCC)}, 2006, pp. 265--284. [Online]. Available:
  \url{http://dx.doi.org/10.1007/11681878\_14}
\BIBentrySTDinterwordspacing

\bibitem{choquette2021label}
C.~A. Choquette-Choo, F.~Tramer, N.~Carlini, and N.~Papernot, ``Label-only
  membership inference attacks,'' in \emph{International Conference on Machine
  Learning}.\hskip 1em plus 0.5em minus 0.4em\relax PMLR, 2021, pp. 1964--1974.

\bibitem{aiprinciples}
\BIBentryALTinterwordspacing
Google. Artificial intelligence at google: Our principles. [Online]. Available:
  \url{https://ai.google/principles}
\BIBentrySTDinterwordspacing

\bibitem{sainath2020streaming}
T.~N. Sainath, Y.~He, B.~Li, A.~Narayanan, R.~Pang, A.~Bruguier, S.-y. Chang,
  W.~Li, R.~Alvarez, Z.~Chen \emph{et~al.}, ``A streaming on-device end-to-end
  model surpassing server-side conventional model quality and latency,'' in
  \emph{ICASSP 2020-2020 IEEE International Conference on Acoustics, Speech and
  Signal Processing (ICASSP)}.\hskip 1em plus 0.5em minus 0.4em\relax IEEE,
  2020, pp. 6059--6063.

\bibitem{huang2022sentence}
W.~R. Huang, C.~Peyser, T.~N. Sainath, R.~Pang, T.~Strohman, and S.~Kumar,
  ``Lookup-table recurrent language models for long tail speech recognition,''
  \emph{arXiv preprint arXiv:2104.04552}, 2021.

\bibitem{kim2017mtr}
C.~Kim, A.~Misra, K.~Chin \emph{et~al.}, ``{Generation of Large-Scale Simulated
  Utterances in Virtual Rooms to Train Deep-Neural Networks for Far-Field
  Speech Recognition in {Google Home}},'' in \emph{Proc. Interspeech}, 2017.

\bibitem{Park2019}
D.~S. Park, W.~Chan, Y.~Zhang, C.~Chiu, B.~Zoph, E.~Cubuk, and Q.~Le,
  ``{SpecAugment: A Simple Data Augmentation Method for Automatic Speech
  Recognition},'' in \emph{Proc. Interspeech}, 2019.

\bibitem{sainath2021cascadedlm}
T.~N. Sainath, Y.~He, A.~Narayanan \emph{et~al.}, ``{An Efficient Streaming
  Non-Recurrent On-Device End-to-End Model with Improvements to Rare-Word
  Modeling},'' in \emph{Proc. of Interspeech}, 2021.

\bibitem{Rami21}
R.~Botros, T.~Sainath, R.~David, E.~Guzman, W.~Li, and Y.~He, ``Tied \& reduced
  rnn-t decoder,'' in \emph{Proc. Interspeech}, 2021.

\bibitem{Vaswani17}
\BIBentryALTinterwordspacing
A.~Vaswani, N.~Shazeer, N.~Parmar, J.~Uszkoreit, L.~Jones, A.~N. Gomez,
  L.~Kaiser, and I.~Polosukhin, ``{Attention Is All You Need},'' \emph{CoRR},
  vol. abs/1706.03762, 2017. [Online]. Available:
  \url{http://arxiv.org/abs/1706.03762}
\BIBentrySTDinterwordspacing

\bibitem{kannan2018analysis}
A.~Kannan, Y.~Wu, P.~Nguyen, T.~N. Sainath, Z.~Chen, and R.~Prabhavalkar, ``An
  analysis of incorporating an external language model into a
  sequence-to-sequence model,'' in \emph{2018 IEEE International Conference on
  Acoustics, Speech and Signal Processing (ICASSP)}.\hskip 1em plus 0.5em minus
  0.4em\relax IEEE, 2018, pp. 1--5828.

\bibitem{variani2020hybrid}
E.~Variani, D.~Rybach, C.~Allauzen, and M.~Riley, ``Hybrid autoregressive
  transducer (hat),'' in \emph{ICASSP 2020-2020 IEEE International Conference
  on Acoustics, Speech and Signal Processing (ICASSP)}.\hskip 1em plus 0.5em
  minus 0.4em\relax IEEE, 2020, pp. 6139--6143.

\bibitem{golovin2017google}
D.~Golovin, B.~Solnik, S.~Moitra, G.~Kochanski, J.~Karro, and D.~Sculley,
  ``Google vizier: A service for black-box optimization,'' in \emph{Proceedings
  of the 23rd ACM SIGKDD international conference on knowledge discovery and
  data mining}, 2017, pp. 1487--1495.

\end{thebibliography}
